\documentclass[11pt,a4paper]{article}
\usepackage[hyperref]{emnlp2018}
\usepackage{times}
\usepackage{latexsym}
\usepackage{url}
\usepackage{graphicx}
\usepackage{amssymb}
\usepackage{amsmath}
\usepackage{multirow}
\usepackage{graphicx}
\usepackage{multirow}
\usepackage{algorithm}
\usepackage{algorithmic}
\usepackage{amssymb}
\usepackage{amsmath}
\usepackage{dsfont}

\aclfinalcopy 
\def\aclpaperid{264} 

\title{Question Generation from SQL Queries Improves \\Neural Semantic Parsing}

\author{Daya Guo$^1$\thanks{\ \ \ Work done while this author was an intern at Microsoft Research.} , Yibo Sun$^3$$^*$, Duyu Tang$^2$, Nan Duan$^2$, Jian Yin$^1$, \\
	\bf Hong Chi$^2$, James Cao$^2$, Peng Chen$^2$, and Ming Zhou$^2$\\
	$^1$ The School of Data and Computer Science, Sun Yat-sen University.\\
	Guangdong Key Laboratory of Big Data Analysis and Processing, Guangzhou, P.R.China\\
	$^2$ Microsoft Research $^3$ Harbin Institute of Technology\\
	{\tt \{guody5@mail2,issjyin@mail\}.sysu.edu.cn}\\
	{\tt \{dutang,nanduan,hongchi,jcao,peche,mingzhou\}@microsoft.com}\\	
	{\tt ybsun@ir.hit.edu.cn}\\
}

\date{}

\begin{document}
\maketitle
\begin{abstract}
	We study how to learn a semantic parser of state-of-the-art accuracy with less supervised training data.
	We conduct our study on WikiSQL, the largest hand-annotated semantic parsing dataset to date.
	First, we demonstrate that question generation is an effective method that empowers us to learn a state-of-the-art neural network based semantic parser with thirty percent of the supervised training data.
	Second, we show that applying question generation to the full supervised training data further improves the state-of-the-art model.
	In addition, we observe that there is a logarithmic relationship between the accuracy of a semantic parser and the amount of training data.
\end{abstract}

\vspace{0.001cm}
\section{Introduction}
Semantic parsing aims to map a natural language utterance to an executable program (logical \mbox{form}) \cite{zelle1996learning,wong2007learning,zettlemoyer2007online}. 
Recently, neural network based approaches \cite{dong-lapata:2016:P16-1,jia-liang:2016:P16-1,xiao-dymetman-gardent:2016:P16-1,guu-EtAl:2017:Long,P18-1069} have achieved promising performance in semantic parsing.
However, neural network approaches are data hungry, which performances closely correlate with the volume of training data.
In this work, we 
study the influence of training data on the accuracy of neural semantic parsing, and how to train a state-of-the-art model with less training data.

We conduct the study on WikiSQL \cite{zhong2017seq2sql}, the largest hand-annotated semantic parsing dataset which is larger than other datasets in terms of both the number of logical forms and the number of schemata.
The task is to map a natural language question to a SQL query. 
We use a state-of-the-art end-to-end semantic parser based on neural \mbox{networks} (detailed in Section \ref{section:semantic-parsing}), and vary the number of supervised training instances. 
Results show that there is a logarithmic relationship between accuracy and the amount of training data, which is consistent with the observations in computer vision tasks \cite{sun2017revisiting}.

We further study how to achieve state-of-the-art parsing accuracy with less supervised data, since annotating a large scale semantic parsing dataset requires funds and domain expertise.
We achieve this through question generation, which generates natural language questions from SQL queries.
Our question generation model is based on sequence-to-sequence learning. 
Latent variables \cite{cao2017latent} are introduced to increase the diversity of generated questions.
The artificially generated question-SQL pairs can be viewed as pseudo-labeled data, which can be combined with a small amount of human-labeled data to train the semantic parser. 

Results on WikiSQL show that  
the state-of-the-art logical form accuracy drops from 60.7\% to 53.7\% with only thirty percent of training data, while increasing to 61.0\% when we combine the pseudo-labeled data generated from the question generation model. 
Applying the question generation model to full training \mbox{data}
brings further improvements with 3.0\% absolute gain.
We further conduct a transfer learning experiment that applies our approach trained on WikiSQL to WikiTableQuestions \cite{pasupat-liang:2015:ACL-IJCNLP}.
Results show that incorporating generated instances improves the state-of-the-art neural semantic parser \cite{krishnamurthy-dasigi-gardner:2017:EMNLP2017}.

\section{Overview of the Approach}
Our task aims to map a question to a SQL query, which is executable over a table to yield the answer. Formally, the task takes a question $q$ and a table $t$ consisting of $n$ column names and $n \times m$ cells as the input, and outputs a SQL query $y$. 
In this section, we describe an overview of our approach, which is composed of several components.

\begin{figure}[h]
	\centering
	\includegraphics[width=.46\textwidth]{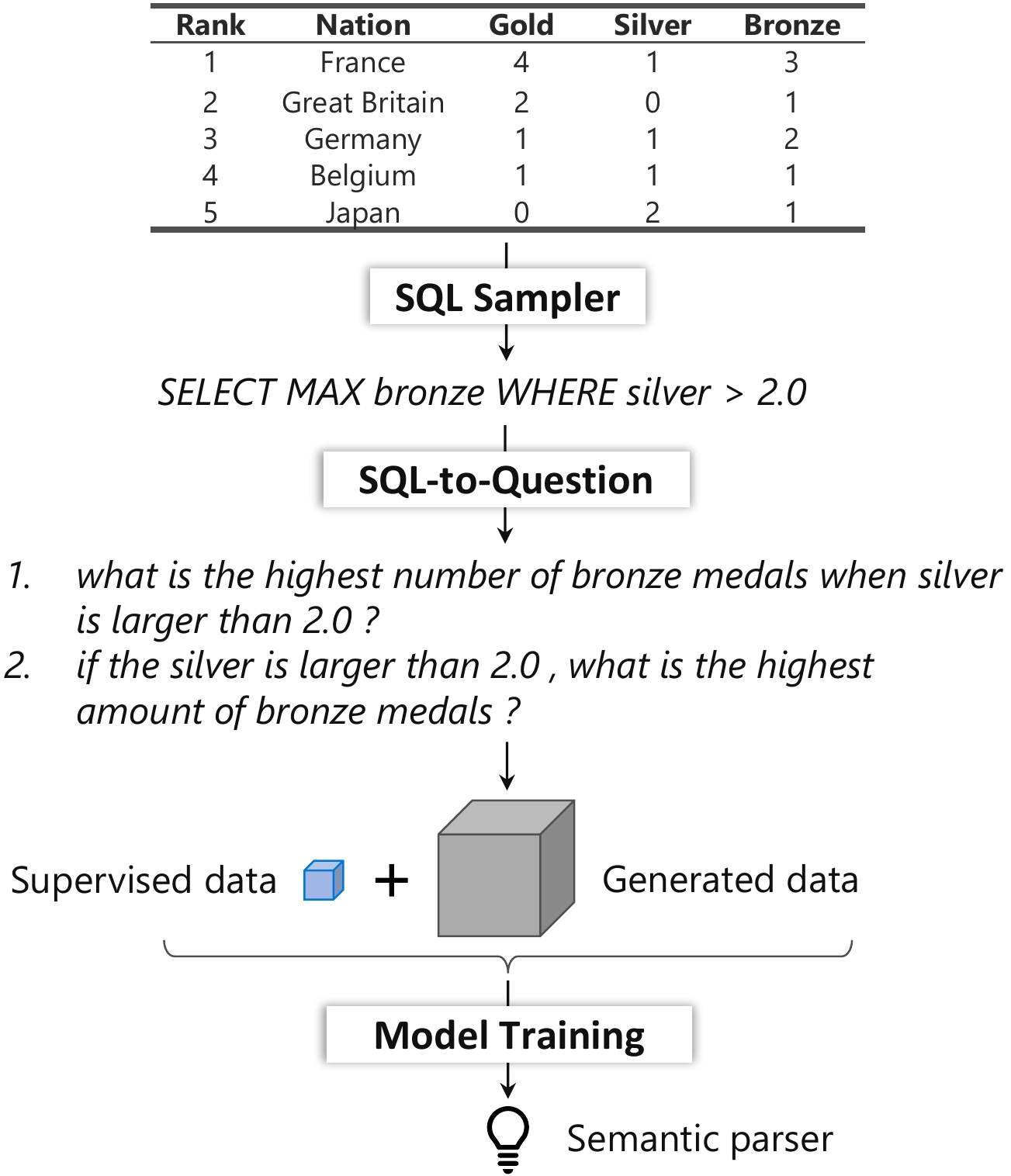}
	\caption{An overview of our approach that improves semantic parsing with question generation.}
	\label{fig:workflow}
\end{figure}
Figure \ref{fig:workflow} gives an overview of our approach. 
First, given a table, a SQL query sampler is used to sample valid, realistic, and representative SQL queries.
Second, a question generation component takes SQL queries as inputs to obtain natural language questions. 
Here, the question generation model is learnt from a small-scale supervised training data that consists of SQL-question pairs. 
Lastly, the generated question-SQL pairs are viewed as the pseudo-labeled data, which are combined with the supervised training data to train the semantic parser.  

Since we conduct the experiment on WikiSQL dataset, we follow \newcite{zhong2017seq2sql} and use the same template-based SQL sampler, as summarized in Table \ref{table:sql-sampler}.
The details about the semantic parser and the question generation model 
will be introduced in Sections \ref{section:semantic-parsing} and Section \ref{section:qg}, respectively.

\begin{table}[h]\centering
	\small
	\begin{tabular}{p{1cm}|p{5.5cm}}
		\hline 
		\multicolumn{2}{c}{\bf Format of a Sampled SQL Query}
\\
		\hline   
		\multicolumn{2}{p{7cm}}{SELECT $agg\_op$ $agg\_col$ From $table$ WHERE $cond1\_col$ $cond1\_op$ $cond1$ AND $cond2\_col$ ... }\\
		\hline	
		\multicolumn{2}{c}{\bf Sampling Rules}\\
		\hline
		Variable & \multicolumn{1}{c}{Sampling range} \\
		\hline
		$agg\_col$ or $cond\_col$& The aggregation column $agg\_col$ and the condition column $cond\_col$ can be one of columns in the table.\\
		\hline
		$agg\_op$ & The aggregation operator $agg\_op$ can be empty or \textit{COUNT}. If the type of $agg\_col$  is numeric, $agg\_op$ can additionally be one of \textit{MAX} and \textit{MIN}. \\
		\hline
		$cond\_op$& The condition operator $cond\_op$ is $=$. If the type of  $cond\_col$ is numeric, $cond\_op$ can additionally be one of $>$ and $<$.\\
		\hline
		$cond$& The condition value $cond$ can be any cell value under the $cond\_col$. If the type of  $cond\_col$ is numeric, $cond$ can be numerical value sampled from minimum value to maximum value in the $cond\_col$.\\
		\hline		
		\multicolumn{2}{c}{\bf Filter Rules}\\
		\hline
		\multicolumn{2}{p{6.5cm}}{		
			1.The condition will be removed if doing the action does not change the execution result.}	\\
		\multicolumn{2}{p{6.5cm}}{		
			2.We only save the sampled SQL queries that produce non-empty result set.}	\\

		\hline		
	\end{tabular}
	\caption{The SQL sampler of \cite{zhong2017seq2sql}.}
	\label{table:sql-sampler}
\end{table}

\section{Semantic Parsing Model}\label{section:semantic-parsing}
We use a state-of-the-art end-to-end semantic parser \cite{sun2018semantic} that takes a natural language question as the input and outputs a SQL query, which is executed on a table to obtain the answer. 
To make the paper self-contained, we briefly describe the approach in this section.

The semantic parser is abbreviated as \mbox{\textbf{STAMP}}, which is short for Syntax- and Table- Aware seMantic Parser. 
Based on the encoder-decoder framework, STAMP takes a question as the input and generates a SQL query. It extends pointer networks \cite{zhong2017seq2sql,vinyals2015pointer} by incorporating three ``channels'' in the decoder, in which the column channel predicts column names, the value channel predicts table cells and the SQL channel predicts SQL keywords.
An additional switching gate selects which channel to be used for generation.
In STAMP, the probability of a token to be generated is calculated as Equation \ref{equa:our}, where $p_z(\cdot)$ is the probability of the channel $z_t$ to be chosen, and $p_w(\cdot)$ is the probability distribution of generating a word $y_t$ from the selected channel.
\vspace{-0.3cm}
\begin{equation}\label{equa:our}
p(y_t| y_{<t}, x) = \sum_{z_t} p_w(y_t | z_t, y_{<t}, x) p_z(z_t| y_{<t}, x)
\end{equation}

Specifically, the encoder takes a question as the input, uses bidirectional RNN with GRU cells to compute the hidden states, and feeds the concatenation of both ends as the initial state of the decoder.
The decoder has another GRU to calculate the hidden states.

Each channel is implemented with an attentional neural network.
In the SQL channel, the input of the attention module includes the decoder hidden state and the embedding of the SQL keyword to be calculated (i.e. $e^{sql}_i$).
\begin{equation}\label{equa:sql-channel}
p_w^{sql}(i)  \propto exp(W_{sql} [h^{dec}_t;e^{sql}_i])
\end{equation}

In the column channel, the vector of a column name includes two parts, as given in Equation \ref{equa:column-channel}.
The first vector ($h^{col}_i$) is calculated with a bidirectional GRU because a column name might contain multiple words. 
The second vector is a question-aware cell vector, which is weighted averaged over the cell vectors belonging to the column.
Cell vectors ($h^{cell}_i$) are also obtained by a bidirectional GRU. The importance of a cell is measured by the number of co-occurred question words, which is further normalized through a $softmax$ function to yield the final weight $\alpha^{cell}_j \in [0,1]$. 
\begin{equation}\label{equa:column-channel}
p_w^{col}(i)\propto exp(W_{col} [h^{dec}_t;h^{col}_i;\sum_{j \in col_i} \alpha^{cell}_j h^{cell}_j])
\end{equation}

In the value channel, the model has two distributions and weighted average them as Equation \ref{equa:cell-channel}.
Similar to $p^{sql}(\cdot)$, a standard cell distribution $\hat{p}_w^{cell}(\cdot)$ is calculated over the cells belonging to the last predicted column name. 
They incorporate an additional probability distribution $\alpha^{cell}(\cdot)$ based on the aforementioned word co-occurrence.
The  hyper parameter $\lambda$ is tuned on the dev set.
\begin{equation}\label{equa:cell-channel}
p_w^{cell}(j) = \lambda \hat{p}_w^{cell}(j) + (1-\lambda)\alpha^{cell}_j
\end{equation} 

Please see more details on model training and inference in \newcite{sun2018semantic}.

\section{Question Generation Model}\label{section:qg}
In this section, we present our SQL-to-question generation approach, which takes a SQL query as the input and outputs a natural language question.
Our approach is  
based on sequence-to-sequence learning \cite{sutskever2014sequence,Bahdanau2015}.
In order to replicate rare words from SQL queries, we adopt the copying mechanism. 
In addition, we incorporate latent variables to increase the diversity of generated questions.

\subsection{Encoder-Decoder}
\paragraph{Encoder:}
A bidirectional RNN with gated recurrent unit (GRU) \cite{cho-EtAl:2014:EMNLP2014} is used as the encoder to read a SQL query $x=(x_1,...,x_T)$. The forward RNN reads a SQL query in a left-to-right direction, obtaining hidden states $(\overrightarrow{h_1},...,\overrightarrow{h_T})$. The backward RNN reads reversely and outputs $(\overleftarrow{h_1},...,\overleftarrow{h_T})$. We then get the final representation $(h_1,...,h_T)$ for each word in the query, where $h_j=[\overrightarrow{h_j};\overleftarrow{h_j}]$. 
The representation of the source sentence $h_x$ $=$ ($[\overrightarrow{h_T};\overleftarrow{h_1}]$) is used as initial hidden state of the decoder.

\paragraph{Decoder:}
We use a GRU with an attention mechanism as the decoder.
At each time-step $t$, the attention mechanism obtains the context vector $c_{t}$ that is computed same as the multiplicative attention \cite{luong2015effective}.
Afterwards, the concatenation of the context vector, the embedding of the previously predicted word $y_{t-1}$, and the last hidden state $s_{t-1}$ is fed to the next step.
\begin{equation}\label{equa:update}
s_t= GRU(s_{t-1},y_{t-1},c_t)
\end{equation}
After obtaining hidden states $s_{t}$, we adopt the copying mechanism that predicts a word from the target vocabulary or from the source sentence (detailed in Subsection \ref{subsec:copy mechanism}). 

\subsection{Incorporating Copying Mechanism}
\label{subsec:copy mechanism}
In our task, the generated question utterances typically include informative yet low-frequency words 
such as named entities or numbers.
Usually, these words are not included in the target vocabulary but  come from SQL queries. 
To address this, we follow {CopyNet} \cite{gu-EtAl:2016:P16-1} and incorporate a copying mechanism to select whether to generate from the vocabulary or copy from SQL queries.

The probability distribution of generating the $t$-th word is calculated as Equation \ref{equa:copy distribution}, where $\psi_g(\cdot)$ and $\psi_c(\cdot)$ are scoring functions for generating from the vocabulary $\boldsymbol{\nu}$ and copying from the source sentence x, respectively.
\begin{equation}\label{equa:copy distribution}
p(y_t|y_{<t},x)=\frac{e^{\psi_g(y_t)}+e^{\psi_c(y_t)}}{\sum_{v\in\boldsymbol{\nu}}e^{\psi_g(v)}+\sum_{w\in{\textbf{x}}}e^{\psi_c(w)}}
\end{equation}

The two scoring functions are calculated as follows, where $W_g$ and $W_c$ are model parameters, $v_i$ is the one-hot indicator vector for $y_i$ and $h_i$ is the hidden state of word $y_i$ in the source sentence.
\begin{equation}\label{equa:generate}
\begin{split}
&\psi_g(y_i)=v_i^TW_gs_t \\
&\psi_c(y_i)=tanh({h_i}^TW_c)s_t
\end{split}
\end{equation}

\subsection{Incorporating Latent Variable}  
Increasing the diversity of generated questions is very important to improve accuracy, generalization, and stability of the semantic parser, since this increases the mount of training data and produces more diverse questions for the same intent. 
In this work, we incorporate stochastic latent variables \cite{cao2017latent,serban2017hierarchical} to the sequence-to-sequence model in order to increase question diversity. 

Specifically, we introduce a latent variable $z\sim p(z)$, which is a standard Gaussian distribution $\mathcal{N}(0, I_n)$ in our case, and calculate the likelihood of a target sentence $y$ as follows:
\begin{equation}\label{equa:likelihood}
p(y|x)= \int_{z} p(y|z,x)p(z)\, dz 
\end{equation}

We maximize the evidence lower bound (ELBO), which decomposes the loss into two parts, including (1) the KL divergence between the posterior distribution and the prior distribution, and (2) a cross-entropy loss between the generated question and the ground truth. 
\begin{align}\label{equa:KL}
\notag
logp(y|x)\geq -D_{KL}(Q(z|x,y)||p(z)) \\
+ E_{z\sim Q}logp(y|z,x)
\end{align}  
The KL divergence in Equation \ref{equa:KL} is calculated as follow, where $n$ is the dimensionality of $z$.
\begin{equation}\label{equa:d-kl}
\begin{split}
&D_{KL}(Q(z|x,y)||p(z))= \\
&-\frac{1}{2}\sum_{j=1}^n(1+log(\sigma_j^2)-\mu_j^2-\sigma_j^2)
\end{split}
\end{equation}  
$Q(z|x,y)$ is a posterior distribution with Gaussian distribution. The mean $\mu$ and standard deviation $\sigma$ are calculated as follows, where $h_x$ and $h_y$ are representations of source and target sentences in the encoder, respectively. Similar to $h_x$, $h_y$ is obtained by encoding the target sentence. 
\begin{equation}\label{equa:mean_variance}
\begin{split}
&\mu=W_\mu[h_x;h_y]+b_\mu \\
&log(\sigma^2)=W_\sigma[h_x;h_y]+b_\sigma
\end{split}
\end{equation}

\subsection{Training and Inference}

At the training phase, we sample $z$ from $Q(z|x,y)$ using the re-parametrization trick \cite{kingma2014auto-encoding}, and concatenate the source last hidden state $h_x$ and $z$ as the initial state of the decoder. 
Since the model tends to ignore the latent variables by forcing the KL divergence to 0 \cite{bowman2016generating}, 
we add a variable weight to the KL term during training. 
At the inference phase, the model will generate different questions by first sampling $z$ from $p(z)$, concatenating $h_x$ and $z$ as the initial state of the decoder, and then decoding deterministically for each sample.

Here, we list our training details. 
We set the dimension of the encoder hidden state as 300, and the dimension of the latent variable $z$ as 64. 
We use dropout with a rate of 0.5, which is applied to the inputs of RNN.
Model parameters are initialized with uniform distribution, and updated with stochastic gradient decent.
Word embedding values are initialized with Glove vectors \cite{pennington-socher-manning:2014:EMNLP2014}.
We set the learning rate as 0.1 and the batch size as 32. 
We tune hyper parameters on the development, and use beam search in the inference process. 

\section{Experiment}
\begin{table*}[t]
	\centering
	\begin{tabular}{l|c|cc|cc}
		\hline
		\multirow{2}{*}{Methods} & \multirow{2}{*}{Training Data}& \multicolumn{2}{c|}{Dev} & \multicolumn{2}{c}{Test} \\
		\cline{3-6}
		& & {Acc$_{lf}$} & {Acc$_{ex}$} & {Acc$_{lf}$} & {Acc$_{ex}$}\\
		\hline
		Attentional Seq2Seq & 100\%& 23.3\%& 37.0\%& 23.4\%& 35.9\% \\
		Aug.PntNet~\cite{zhong2017seq2sql} & 100\% &44.1\%& 53.8\%& 43.3\%& 53.3\%\\
		Aug.PntNet~(re-implemented by us) & 100\%& 51.5\%& 58.9\%& 52.1\%& 59.2\% \\
		Seq2SQL~\cite{zhong2017seq2sql} & 100\%& 49.5\%& 60.8\%& 48.3\%& 59.4\%\\
		SQLNet~\cite{xu2017sqlnet} & 100\%& -- & 69.8\% & -- & 68.0\%\\
		\hline
		STAMP & 30\%& 54.6\%& 69.7\%& 53.7\%& 68.9\%\\
		STAMP + QG & 30\%& 61.6\%& 74.4\%& 61.2\% & 73.9\%\\
		STAMP   & 100\%& 61.5\%& 74.8\%& 60.7\%& 74.4\% \\
		STAMP + QG & 100\%& 64.3\%&  76.5\%& 63.7\%&  75.5\%\\
		\hline
	\end{tabular}
	\caption{Performance of different approaches on the WikiSQL dataset. The two evaluation metrics are logical form accuracy (Acc$_{lf}$) and execution accuracy (Acc$_{ex}$). The settings of the training data represent the proportion of supervised data we use. }
	\label{table:compare-to-other-alg}
\end{table*}

We conduct experiments on the WikiSQL  dataset\footnote{\url{https://github.com/salesforce/WikiSQL}} \cite{zhong2017seq2sql}. 
WikiSQL is the largest hand-annotated semantic parsing dataset which is an order of magnitude larger than other datasets in terms of both the number of logical forms and the number of schemata (tables).
WikiSQL is built by crowd-sourcing on Amazon Mechanical Turk, including 61,297 examples for training, and 9,145/17,284 examples for development/testing. 
Each instance consists of a natural language question, a SQL query, a table and a result. 
Here, we follow \newcite{zhong2017seq2sql} to use two evaluation metrics. One is logical form accuracy (Acc$_{lf}$), which measures the percentage of exact string match between the generated SQL queries and the ground truth SQL queries. 
Since different logical forms might obtain the same result, another metric is execution accuracy (Acc$_{ex}$), which is the percentage of the generated SQL queries that result in the correct answer.

\subsection{Impact of Data Size}
We study how the number of training instances affects the accuracy of semantic parsing.

\begin{figure}[h]
	\centering
	\includegraphics[width=.48\textwidth]{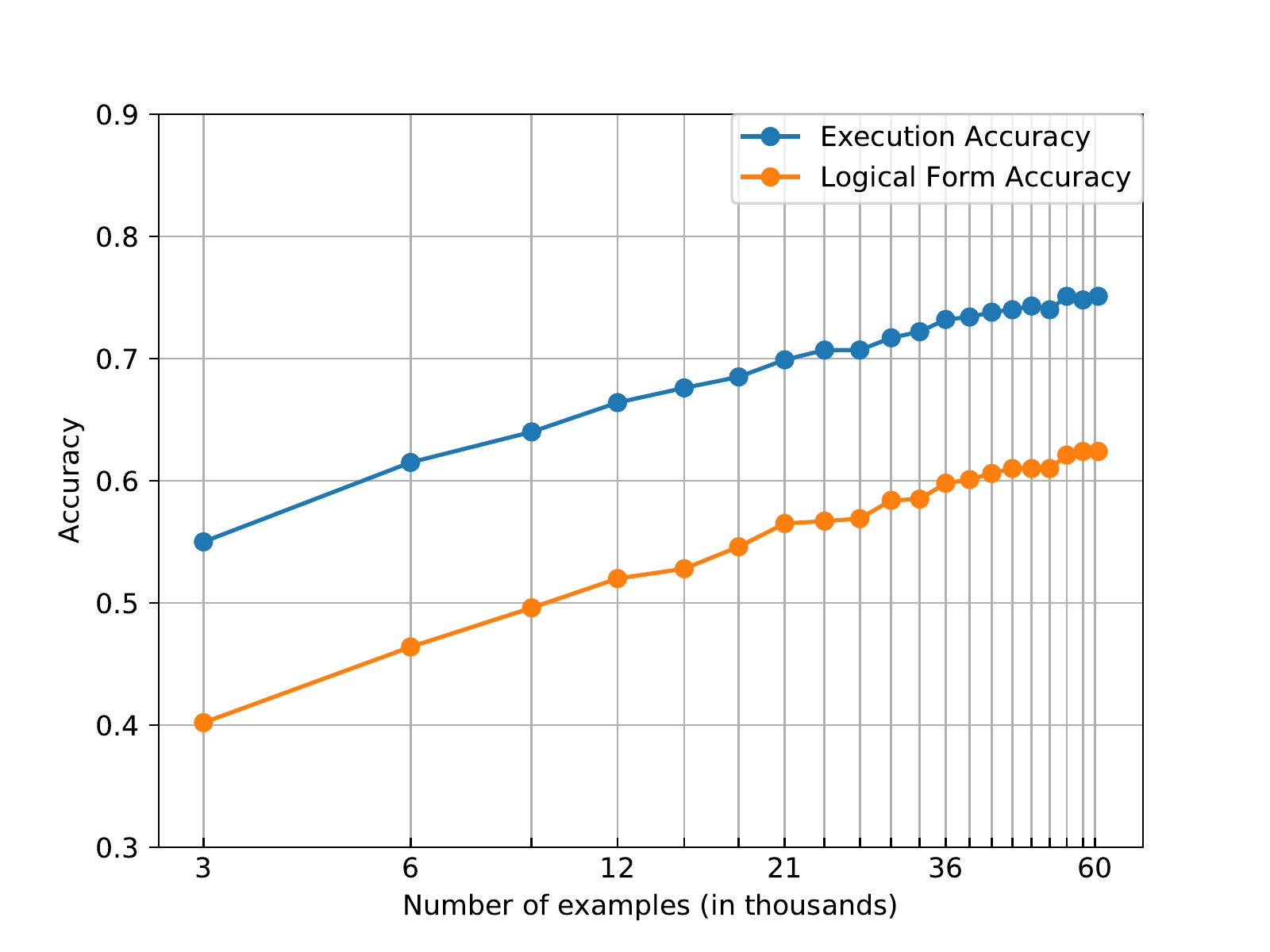}
	\caption{Semantic parsing accuracies of the \mbox{STAMP} model on WikiSQL. The $x$-axis is the training data size in log-scale, and the $y$-axis includes two evaluation metrics Acc$_{lf}$ and Acc$_{ex}$.}
	\label{fig:log}
\end{figure}

In this experiment, we randomly sample 20 subsets of examples from the WikiSQL training data, incrementally increased by 3K examples (about 1/20 of the full WikiSQL training data).
We use the same training protocol and report the accuracy of the STAMP model on the dev set.
Results are given in Figure \ref{fig:log}. It is not surprising that more training examples bring higher accuracy. 
Interestingly, we observe that both accuracies of the neural network based semantic parser grow logarithmically as training data expands, which is consistent with the observations in computer vision tasks \cite{sun2017revisiting}.

\subsection{Model Comparisons}
We report the results of existing methods on WikiSQL, and demonstrate that question generation is an effective way to improve the accuracy of semantic parsing. 
\newcite{zhong2017seq2sql} implement several methods, including \textbf{Attentional Seq2Seq}, which is a basic attentional sequence-to-sequence learning baseline;
\textbf{Aug.PntNet}, which is an augmented pointer network in which words of the target sequence come from the source sequence;
and \textbf{Seq2SQL} which extends Aug.PntNet by further learning two separate classifiers for SELECT aggregator and SELECT column.
\newcite{xu2017sqlnet} develop \textbf{SQLNet}, which uses two separate models to predict SELECT and WHERE clauses, respectively, and introduce a sequence-to-set neural network to predict the WHERE clause.~\textbf{STAMP}  stands for the semantic parser which has been described in Section \ref{section:semantic-parsing}.

\begin{figure}[h]
	\centering
	\includegraphics[width=.48\textwidth]{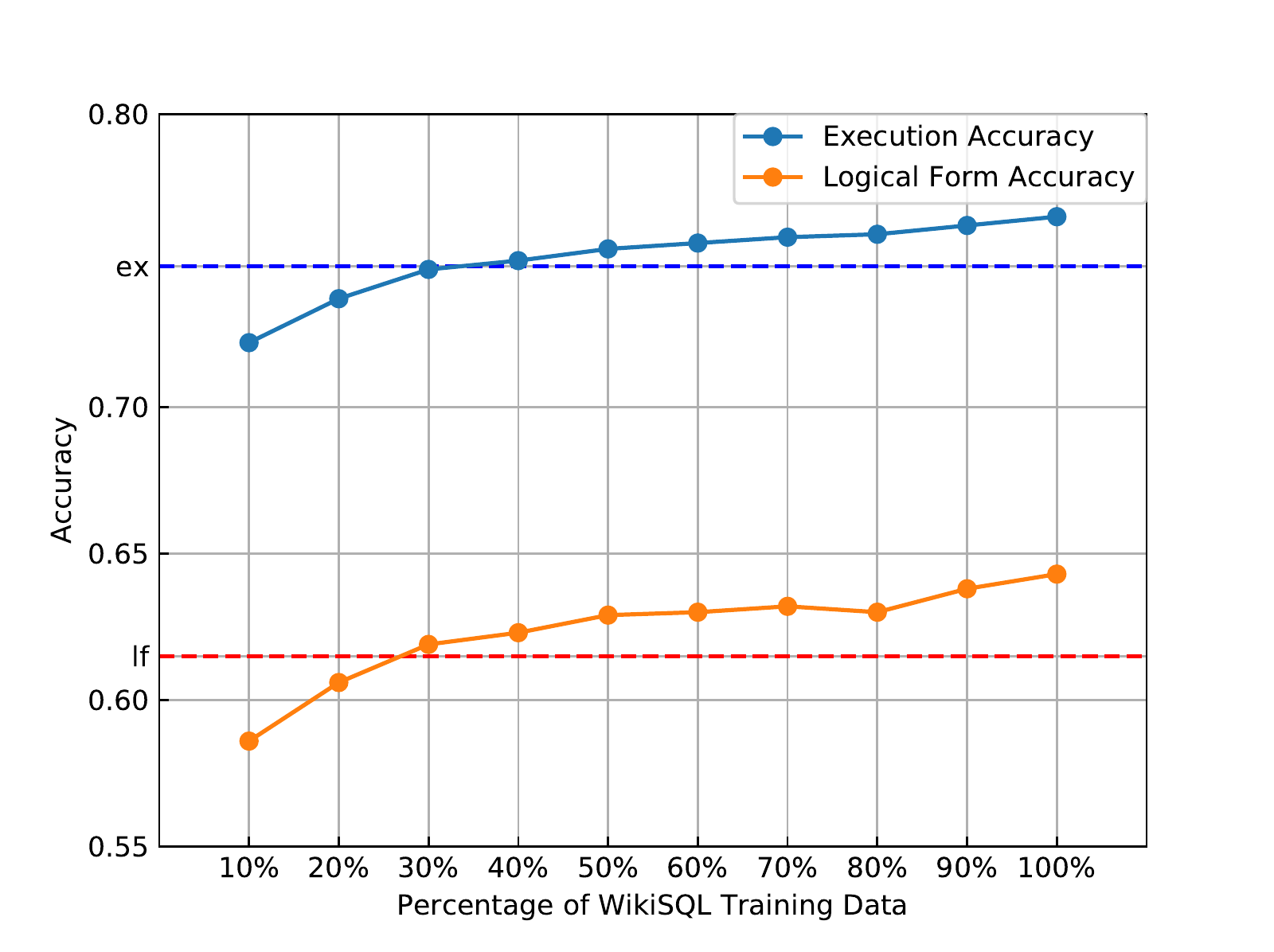}
	\caption{Accuracies of STAMP+QG with different portions of supervised data. Dashed lines are Acc$_{lf}$ and Acc$_{ex}$  of STAMP on the full training data.}
	\label{fig:ac}
\end{figure}

From Table \ref{table:compare-to-other-alg}, we can see that 
STAMP \mbox{performs} better than existing systems when trained on the full WikiSQL training dataset, achieving  state-of-the-art execution accuracy and logical form accuracy on WikiSQL.
We further conduct experiments to demonstrate the effectiveness of our question generation driven approach.
We run the entire pipeline (STAMP+QG) with different percentages of training data. 
The second column ``Training Data'' in Table \ref{table:compare-to-other-alg} and the $x$-axis in Figure \ref{fig:ac} represent the proportion of WikiSQL training data we use for training the QG model and semantic parser. 
That is to say, STAMP +QG with 30\% means that we sample 30\% WikiSQL training data to train the QG model, and then combine QG generated data and exactly the same 30\% WikiSQL training data we sampled before to train the semantic parser. In this experiment, we sample five SQL queries for each table in the training data, resulting in 43.5K SQL queries. Applying the QG model on these SQL queries, we get 92.8K SQL-question pairs. 
From Figure \ref{fig:ac}, we see that accuracy increases as the amount of supervised training data expands.
Results show that QG empowers the STAMP model to achieve the same accuracy on WikiSQL dataset with 30\% of the training data. Applying QG to the STAMP model under the full setting brings further improvements, resulting in new state-of-the-art accuracies.

\begin{table*}[t]
	\centering
	\begin{tabular}{l|ccc|ccc}
		\hline
		\multirow{2}{*}{Methods} & \multicolumn{3}{c|}{Dev} & \multicolumn{3}{c}{Test} \\
		\cline{2-7}
		& {Acc$_{sel}$} & {Acc$_{agg}$} & Acc$_{where}$ & {Acc$_{sel}$} & {Acc$_{agg}$}& Acc$_{where}$\\
		\hline
		Aug.PntNet (re-implemented by us)& 80.9\%& 89.3\%& 62.1\%& 81.3\%& 89.7\%& 62.1\%\\
		Seq2SQL~\cite{zhong2017seq2sql} & 89.6\%& 90.0\%& 62.1\%& 88.9\%& 90.1\%& 60.2\%\\
		SQLNet~\cite{xu2017sqlnet} & 91.5\% & 90.1\%& 74.1\%& 90.9\% & 90.3\%& 71.9\%\\
		\hline
		STAMP & 89.4\%& 89.5\%& 77.1\%& 88.9\%& 89.7\%& 76.0\%\\
		STAMP+QG& 89.7\%& 90.1\%& 79.8\%& 89.1\%& 90.2\%& 79.0\%\\
		\hline
	\end{tabular}
	\caption{Fine-grained accuracies on the WikiSQL dev and test sets. Logical form accuracy (Acc$_{lf}$) is evaluated on SELECT column (Acc$_{sel}$) , SELECT aggregator (Acc$_{agg}$), and WHERE clause (Acc$_{where}$), respectively. All these models are trained on the full WikiSQL training data.}
	\label{table:fine-grained-results}
\end{table*}
\subsection{Fine-grained Accuracies}

Since SQL queries in WikiSQL consist of SELECT column, SELECT aggregator, and WHERE clause, we report fine-grained accuracies with regard to these aspects, respectively.

From Table \ref{table:fine-grained-results}, we observe that the main advantage of STAMP+QG over STAMP comes from the prediction of the WHERE clause, which is also the main challenge of the WikiSQL dataset.
We further analyze STAMP and STAMP+QG on the WHERE clause by splitting the dev and test sets into three groups according to the number of \mbox{conditions} in the WHERE clause. From Table \ref{table:difficulty}, we see that combining QG is helpful when the number of WHERE conditions is more than one. 
The main reason is that dominant instances in the WikiSQL training set have only 
one WHERE condition, as shown in Table \ref{table:distribution}, thus the model might not have memorized enough patterns for the other two limited-data groups.
Therefore, the pseudo-labeled instances generated by our  
SQL sampler and QG approach  
are more precious to the limited-data groups (i.e \#where =2 and \#where$\geq$3).

\begin{table}[h]
	\centering
	\begin{tabular}{c|cc|cc}
		\hline
		\multirow{2}{*}{\#where} & \multicolumn{2}{c|}{STAMP} & \multicolumn{2}{c}{STAMP+QG} \\
		\cline{2-5}
		& {dev} & {test} & {dev} & {test} \\
		\hline
		$=$ 1 & 80.9\%& 80.2\%& 81.5\%& 80.9\%\\
		$=$ 2 & 65.1\%& 65.4\%& 68.3\%& 66.9\%\\
		$\geq$ 3 & 44.1\%& 48.2\%& 53.4\%& 51.9\%\\
		\hline
	\end{tabular}
	\caption{Execution accuracy (Acc$_{ex}$) on different groups of WikiSQL dev and test sets.}
	\label{table:difficulty}
\end{table}

\begin{table}[h]
	\centering
	\begin{tabular}{c|c|c}
		\hline
		\#where& supervised data & generated data\\
		\hline
		$=$ 1 & 69.1\%& 55.4\%\\
		$=$ 2  & 24.1\%& 33.0\%\\
		$\geq$ 3 & 6.1\%& 11.4\%\\
		\hline
	\end{tabular}
	\caption{Distribution of the number of WHERE conditions in supervised and generated data. }
	\label{table:distribution}
\end{table}

\subsection{Influences of Different QG Variations}

To better understand how various components in our QG model impact the overall performance, we study different QG model variations.
We use three evaluation metrics, including two accuracies and BLEU score \cite{papineni2002bleu}. The BLEU score evaluates the question generation.
\begin{table}[h]
	\centering
	\begin{tabular}{l|c|ccc}
		\hline
		Methods & Scale& {BLEU} & {Acc$_{lf}$} & Acc$_{ex}$ \\
		\hline
		s2s& 30\%& 20.6& 59.0\%& 72.1\%\\
		s2s+lv& 30\%& 22.1& 60.0\%& 72.3\%\\
		s2s+cp& 30\%& 29.6& 60.8\%& 73.5\%\\
		s2s+cp+lv  &30\% & 29.5& 61.2\%& 73.9\%\\
		\hline
		s2s& 100\%&  26.0& 62.6\%& 74.9\%\\
		s2s+lv& 100\%&  26.3& 63.0\%& 75.3\%\\
		s2s+cp& 100\%&  31.5& 63.2\%& 75.6\%\\
		s2s+cp+lv &100\% &  31.6& 63.7\%& 75.5\%\\
		\hline
	\end{tabular}
	\caption{Performances of different question generation variations.}
	\label{table:QG ablation studies}
\end{table}
\begin{table*}[t]
	\centering
	\begin{tabular}{l|l}
		\hline
		SQL& SELECT COUNT 2nd leg WHERE aggregate = 7-2\\
		\hline
		Question (ground truth) & what is the total number of 2nd leg where aggregate is 7-2\\
		\hline
		Question (s2s + cp)  &how many 2nd leg with aggregate being 7-2
		\\\hline
		\multirow{3}{*}{Question (s2s + cp + lv) } & (1) what is the total number of 2nd leg when the aggregate is 7-2 ?
		\\
		& (2) how many 2nd leg with aggregate being 7-2
		\\
		& (3) name the number of 2nd leg for 7-2
		\\		
		\hline
	\end{tabular}
	\caption{Generated examples from different question generation model variations.}
	\label{table:QG example}
\end{table*}

Results are shown in Table \ref{table:QG ablation studies}, in which 
\textbf{s2s} represents the basic attentional sequence-to-sequence learning model \cite{luong2015effective}, \textbf{cp} means the copying mechanism, and \textbf{lv} stands for the latent variable. 
We can see that 
incorporating a latent variable improves QG model performance, especially in limit-supervision scenarios. This is consistent with our intuition that  the performance of the QG model is improved by incorporating the copying mechanism, since rare words of great importance mainly come from the input sequence.

To better understand the impact of incorporating a latent variable, we show examples generated by different QG variations in Table \ref{table:QG example}. We can see that incorporating a latent variable empowers the model to generate diverse questions for the same intent.

\subsection{Transfer Learning on WikiTableQuestions}

In this part, we conduct an extensional experiment on WikiTableQuestions\footnote{\url{https://nlp.stanford.edu/software/sempre/wikitable/}} \cite{pasupat-liang:2015:ACL-IJCNLP} in a transfer learning scenario to verify the effectiveness of our approach. 
WikiTableQuestions contains 22,033 complex questions on 2,108 Wikipedia tables. Each instance consists of a natural language question, a table and an answer. 
Following \newcite{pasupat-liang:2015:ACL-IJCNLP}, we report development accuracy which is averaged over the first three 80-20 training data splits. Test accuracy is reported on the train-test data.

In this experiment, we apply the QG model learnt from WikiSQL to improve the state-of-the-art semantic parser  \cite{krishnamurthy-dasigi-gardner:2017:EMNLP2017} on this dataset. 
Different from WikiSQL, this dataset requires question-answer pairs for training. 
Thus, we generate question-answer pairs by follow steps. 
We first sample SQL queries on the tables from WikiTableQuestions, and then use our QG model to generate question-SQL pairs.
Afterwards, we obtain question-answer pairs by executing SQL queries. 
The generated question-answer pairs will be combined with the original WikiTableQuestions training data to train the model.

\begin{table}[h]
	\centering
	\begin{tabular}{l|cc}
		\hline
		& {Dev} & {Test} \\
		\hline
		\newcite{pasupat-liang:2015:ACL-IJCNLP} & 37.0\%& 37.1\%\\
		\newcite{neelakantan2016learning}& 37.5\%& 37.7\%\\
		\newcite{haug2017neural} &- & 38.7\%\\
		\newcite{zhang-pasupat-liang:2017:EMNLP2017} & 40.4\%& 43.7\%\\
		\hline
		STAMP (WikiSQL) & -& 14.5\%\\
		STAMP (WikiSQL) + QG &-& 15.2\%\\
		NSP & 41.9\%& 43.8\%\\
		NSP + QG & 42.2\%& 44.2\%\\
		\hline
	\end{tabular}
	\caption{Accuracy (Acc$_{ex}$) of different approaches on WikiTableQuestion dev and test sets.}
	\label{table:wikitablequestion}
\end{table}

Results are shown in Table \ref{table:wikitablequestion}, in which \textbf{NSP} is short for the state-of-the-art neural semantic parser \cite{krishnamurthy-dasigi-gardner:2017:EMNLP2017}. Since the train-test data used in {NSP} is different from others, we retrain the {NSP} under the same protocol. 
\textbf{STAMP (WikiSQL)} means that the {STAMP} model trained on WikiSQL is directly tested on WikiTableQuestions. 
Despite applying QG slightly improves STAMP in this setting, the low accuracy reflects the different question distribution between these two datasets.
In the supervised learning setting, we can see that incorporating QG further improves the accuracy of {NSP} from 43.8\% to 44.2\%.

\subsection{Discussion}
To better understand the limitations of our QG model, we analyze a randomly selected set of 100 questions. 
We observe that 27\% examples do not correctly express the meanings of SQL queries, among which the majority of them miss information from the WHERE clause. 
This problem might be mitigated by incorporating a dedicated encoder/decoder that takes into account the SQL structure.
Among the other 73\% of examples that correctly express SQL queries, there are two potential directions to make further improvements.
The first direction is to leverage table information such as the type of a column name or column-cell correlations. 
For instance, without knowing that \mbox{cells} under the column name ``\textit{built}'' are all building years, the model hardly predicts a question ``\textit{what is the average building year for superb?}'' for ``\textit{\mbox{SELECT} AVG built WHERE name = superb}''.
The second direction is to incorporate common knowledge, which would help the model to predict \textit{the earliest week} rather than
\textit{the lowest week}.

\section{Related Work}
\paragraph{Semantic Parsing.}
Semantic parsing
is a fundamental problem in NLP that maps natural language utterances to logical forms, which could be executed to obtain the answer (denotation) \cite{Zettlemoyer05,liang2011learning,berant2013semantic,krishnamurthy2013jointly,pasupat-liang:2016:P16-1,iyer-EtAl:2017:Long}.
Existing works can be classified into three areas, including
(1) the language of the logical form, e.g. first-order logic, lambda calculus, lambda dependency-based compositional semantics (lambda DCS) and structured query language (SQL);
(2) the form of the knowledge base, e.g. facts from large collaborative knowledge bases, semi-structured tables and images;
and (3) the supervision used for learning the semantic parser, e.g. question-denotation pairs and question-logical form pairs.
In this work, we regard the table as the knowledge base,
which is critical for accessing relational databases with natural language, and also for serving information retrieval for structured data.
We use SQL as the logical form, which has a broad acceptance to the public.
In terms of supervision, this work 
uses a small portion of question-logical form pairs to initialize the QA model and train the QG model, and incorporate more generated question-logical form pairs to further improve the QA model.

\paragraph{Question Generation}
Our work also relates to the area of question generation, which has drawn plenty of attention recently partly influenced by the remarkable success of neural networks in text generation.
Studies in this area are classified based on the definition of the answer, including a sentence \cite{heilman2011automatic}, a topic word  \cite{chali2015towards}, a fact (including a subject, a relation phrase and an object) from knowledge bases \cite{serban-EtAl:2016:P16-1}, an image \cite{mostafazadeh2016generating}, etc.
Recent studies in machine reading comprehension generate questions from an answer span and its context from the document \cite{du-shao-cardie:2017:Long,golub-EtAl:2017:EMNLP2017}.
\newcite{wang2015building} first generate logical forms, and then use AMTurkers to paraphrase them to get natural language questions. 
\newcite{iyer-EtAl:2017:Long} use a template-based approach based on the Paraphrase Database \cite{ganitkevitch2013ppdb} to generate questions from SQL.
In this work, we generate questions from logical forms, in which the amount of information from two directions are almost identical.
This differs from the majority of existing studies because a question typically conveys less semantic information than the answer.

\paragraph{Improving QA with QG}
This work also relates to recent studies that uses a QG model to improve the performance of a discriminative QA model \cite{wang2017irgan,yang2017semi,duan-EtAl:2017:EMNLP2017,konstas-EtAl:2017:Long}.
The majority of these works generate a question from an answer, while there also exists a recent work \cite{dong2017learning} that generates a question from a question through paraphrasing.
In addition, \newcite{tang2017question} consider QA and QG as dual tasks, and further improve the QG model in a dual learning framework.
These works fall into three categories: (1) regarding the artificially generated results as additional training instances  \cite{yang2017semi,golub-EtAl:2017:EMNLP2017}; (2) using generated questions to calculate additional features \cite{duan-EtAl:2017:EMNLP2017,dong2017learning};
and (3) using the QG results as additional constraints in the training objectives \cite{tang2017question}.
This work belongs to the first direction.
Our QG approach takes a logical form as the input, and considers the diversity of \mbox{generated} questions by incorporating latent variables.

\section{Conclusion}
In this paper, we observe the logarithmic relationship between the accuracy of a semantic parser and the amount of training data, and present an approach that improves neural semantic parsing with question generation. 
We show that question generation helps us obtain a state-of-the-art neural semantic parser with less supervised data, and further improves the state-of-the-art model with \mbox{full} annotated data on WikiSQL and WikiTableQuesions datasets. 
In future work, we would like to make use of table information and external knowledge to improve our QG model. We also plan to apply the approach to other tasks.

\section*{Acknowledgments}

This work is supported by the National Natural Science Foundation of China (61472453, U1401256, U1501252, U1611264,U1711261,U1711262). Thanks to the anonymous reviewers for their helpful comments and suggestions.

\bibliography{emnlp2018}
\bibliographystyle{acl_natbib_nourl}

\end{document}